\documentclass[11pt,a4paper]{article}
\usepackage[hyperref]{acl2020}
\usepackage{times}
\usepackage{latexsym}

\usepackage{amsfonts,amsmath}
\usepackage{url}
\usepackage{paralist}
\usepackage{amsmath}
\usepackage{multirow}
\usepackage{multicol}
\usepackage{caption}
\usepackage{enumerate}
\usepackage{graphicx}
\usepackage{enumitem}
\usepackage{bm}
\usepackage{url}
\usepackage[normalem]{ulem}
\usepackage{subfigure}
\usepackage{mathrsfs}

\aclfinalcopy %  Enter the acl Paper ID here

\title{Continuity of Topic, Interaction, and Query: \\Learning to Quote in Online Conversations}

\author{Lingzhi Wang$^{1,2}$, Jing Li$^3$, Xingshan Zeng$^{1,2}$\thanks{~~Xingshan Zeng is the corresponding author.}, Haisong Zhang$^4$, Kam-Fai Wong$^{1,2}$\\
  $^1$The Chinese University of Hong Kong, Hong Kong, China\\
  $^2$MoE Key Laboratory of High Confidence Software Technologies, China\\
  $^3$Department of Computing, The Hong Kong Polytechnic University, Hong Kong, China\\
  $^4$Tencent AI Lab, China \\
  \tt $^{1,2}$\{lzwang,xszeng,kfwong\}@se.cuhk.edu.hk \\
  \tt $^3$jing-amelia.li@polyu.edu.hk,
  \tt $^4$hansonzhang@tencent.com \\
}

\date{}

\begin{document}
\maketitle

\begin{abstract}
	Quotations are crucial for successful explanations and persuasions in interpersonal communications. 
	However, finding what to quote in a conversation is challenging for both humans and machines. 
	This work studies automatic quotation generation in an online conversation and explores how \emph{language consistency} affects whether a quotation fits the given context.
	Here, we capture the contextual consistency of a quotation in terms of latent topics, interactions with the dialogue history, and coherence to the query turn's existing content.
	Further, an encoder-decoder neural framework is employed to  continue the context with a quotation via language generation. 
	Experiment results on two large-scale datasets in English and Chinese demonstrate that our quotation generation model outperforms the state-of-the-art models.
	Further analysis shows that topic, interaction, and query consistency are all helpful to learn how to quote in online conversations.
	
\end{abstract}
\section{Introduction}

%%% Why we need to generate quotation in conversations

Quotations, or quotes, are memorable phrases or sentences widely echoed to spread patterns of wisdom~\cite{DBLP:conf/naacl/BootenH16}. They are derived from the ancient art of rhetoric and now appearing in various daily activities, ranging from formal writings~\cite{tan2015learning} to everyday conversations~\cite{Lee:2016:QRD:2911451.2914734}, all help us present clear, beautiful, and persuasive language.
However, for many individuals, writing a suitable quotation that fits the ongoing contexts is a daunting task. 
The issue becomes more pressing for quoting in online conversations where quick responses are usually needed on mobile devices~\cite{Lee:2016:QRD:2911451.2914734}.

\begin{figure}[t]
    \centering
    \begin{tabular}{|p{7.2cm}|}
    \hline
    % \\
    $[T1]$: Save your money.  Scuf is the biggest ripoff in gaming. \\
    $[T2]$: What would you suggest instead? \\
    $[T3]$: Just use a normal controller. \\
    $[T4]$: Ooooooh, I get it now...you're just dumb. \\
    $[Q]$: The dumb ones are the people spending over \$100 for a controller. \textcolor{blue}{\textit{A fool and his money are soon parted.}} \\
    % ...\\
    \hline
    \end{tabular}
    \caption{A Reddit conversation snippet about buying a Scuf controller. The quotation is in blue and italic. $[T1]$ to $[T4]$ are history turns while $[Q]$ is for query turn. 
    }
    \label{fig:intro-example}
\end{figure}

%% Our task
To help online users find what to quote in the discussions they are involved in, our work studies how to recommend an ongoing conversation with a quote and ensure its continuity of senses with the existing contexts. 
For task illustration,  Figure \ref{fig:intro-example} displays a Reddit conversation snippet centered around the worthiness to buy a Scuf controller.
To argue against $T4$'s viewpoint, we see the query turn quotes \textit{Tusser}'s old saying for showing that buying a controller is a waste of money.
As can be observed, it is important for a quotation recommendation model to capture the key points being discussed (reflected by words like ``money'' and ``dumb'' here) and align them to words in the quotation to be predicted (such as ``fool'' and ``money''), which allows to quote something relevant and consistent to the previous concern.

%% Our idea

To predict quotations, our work explores semantic consistency of what will be quoted and what was given in the contexts. 
In context modeling, we distinguish the query turn (henceforth \textbf{query}) and the other turns in earlier history (henceforth \textbf{history}), where topic, interaction, and query consistency work together to determine whether a quote fits the contexts.
Here \textbf{topic consistency} ensures that the words in quotation reflect the discussion topic (such as ``fool'' and ``money'' in Figure \ref{fig:intro-example}). \textbf{Interaction consistency} is to identify the turns in history to which the query responds (e.g., $T1$ and $T4$ in Figure \ref{fig:intro-example}) and guide the quote to follow such interaction. \textbf{Query consistency} measures the language coherence of quote in continuing the story started by the query. For example, the quote in Figure \ref{fig:intro-example} is to support the query's argument.

In previous work of quotation recommendation, there are many methods designed for formal writings~\cite{tan2015learning,liu-etal-2019-neural-based}; whereas much fewer efforts are made for online conversations with informal language and complex interactions in their contexts.
\citet{Lee:2016:QRD:2911451.2914734} use a ranking model to recommend quotes for Twitter conversations.
Different from them, we attempt to generate quotations in a word-by-word manner, which allows the semantic consistency of quotes and contexts to be explored.

%% Our method

Concretely, we propose a neural encoder-decoder framework to predict a quotation that continues the given conversation contexts. 
We capture topic consistency with latent topics (i.e., word distributions), which are learned by a neural topic model~\cite{zeng2018topic} and inferred jointly with the other components.
Interaction consistency is modeled with a turn-based attention over the history turns, and the query is additionally encoded to initialize the decoder's states for query consistency.
To the best of our knowledge, we are the first to explore quotation generation in conversations and extensively study the effects of topic, interaction, and query consistency on this task.

%% Our experiment

Our empirical study is conducted on two large-scale datasets, one in Chinese from Weibo and the other in English from Reddit, both of which are constructed as part of this work. 
Experiment results show that our model significantly outperforms both the state-of-the-art model based on quote rankings~\cite{Lee:2016:QRD:2911451.2914734} and the recent topic-aware encoder-decoder model for social media language generation~\cite{wang-etal-2019-topic-aware}. 
For example, we achieve $27.2$ precision@1 on Weibo compared with $24.0$ by \citet{wang-etal-2019-topic-aware}.
Further discussions show that topic, interaction, and query consistency can all usefully indicate what to quote in online conversations.
We also study how length of history and quotation affects the quoting results and find that we perform consistently better than comparison models in varying scenarios.

\section{Related Work}

Our work is in the line with content-based recommendation~\cite{liu-etal-2019-neural-based} or cloze-style reading comprehension~\cite{DBLP:conf/acl/ZhengHS19}, which learns to put suitable text fragments (e.g., words, phrases, sentences) in the given contexts.
Most prior studies explore the task in formal writings, such as citing previous work in scientific papers~\cite{he2010context}, quoting famous sayings in books~\cite{tan2015learning,Tan:2016:NNA:2983323.2983788}, and using idioms in news articles~\cite{liu-etal-2019-neural-based,DBLP:conf/acl/ZhengHS19}.
The language they face is mostly formal and well-edited, while we tackle online conversations exhibiting noisy contexts and hence involving quote consistency modeling with turn interactions.
\citet{Lee:2016:QRD:2911451.2914734} also recommend quotations for conversations.
However, they consider quotations as discrete attributes (for learning to rank) and hence largely ignore the rich information reflected by a quotation's internal word patterns. 
Compared with them, our model learns to quote with language generation, which can usefully exploit how words appear in both contexts and quotations.

For methodology, we are inspired by the encoder-decoder neural language generation models~\cite{sutskever2014sequence, bahdanau2014neural}.
In dialogue domains, such models have achieved huge success in digesting contexts and generate microblog hashtags~\cite{DBLP:conf/naacl/WangLKLS19}, meeting summaries~\cite{DBLP:conf/acl/LiZJR19}, dialogue responses~\cite{DBLP:conf/ijcai/HuCL0MY19}, etc.
Here we explore how the encoder-decoder architecture works to generate quotations in conversations, which has never been studied in existing work.
Our study is also related to previous research to understand conversation contexts~\cite{ma-etal-2018-challenging, liu-chen-2019-reading, DBLP:journals/corr/abs-1902-00164}, where it is shown to be useful to capture interaction structures~\cite{liu-chen-2019-reading} and latent topics~\cite{DBLP:journals/tacl/ZengLHGLK19}. 
For latent topics, we are benefited from the recent advance of neural topic models~\cite{pmlr-v70-miao17a,wang-etal-2019-topic-aware}), which allows end-to-end topic inference in neural architectures.
Nevertheless, none of the above work attempts to study the semantic consistency of quotes in conversation contexts, which is a gap our work fills in.

\section{Our Quotation Generation Model}

\begin{figure}[t]
\centering
\includegraphics[width=0.44\textwidth]{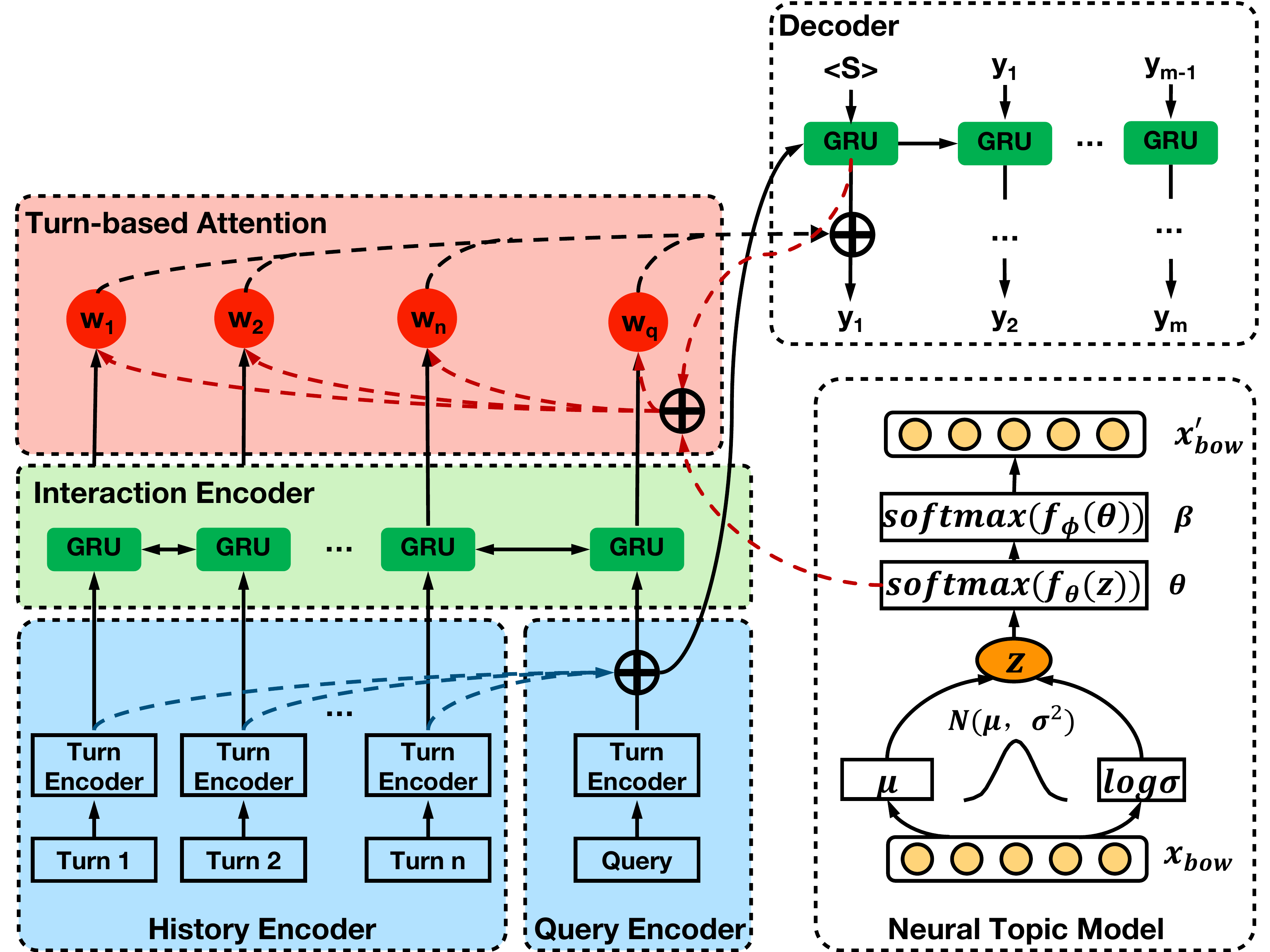}
\caption{\label{fig:sketch} Our encoder-decoder framework for conversation quotation generation. It encodes turn interactions (for both query and earlier history) and latent topics in contexts. The decoder predicts quotes in aware of topic, interaction, and query consistency.}
\end{figure}

This section describes our neural encoder-decoder framework that generates quotations in conversations, whose architecture is shown in Figure~\ref{fig:sketch}. 
The encoding process works for context modeling of turn interactions (described in Section~\ref{ssec:model:encoder}) and latent topics (presented in Section~\ref{ssec:model:topic-model}).
For the decoding process to be discussed in Section \ref{ssec:model:decoder}, we predict words in quotes taking topic, interaction, and query consistency into consideration.
The learning objective of the entire framework will be given at last in Section \ref{ssec:model:objective}.

%In the following, we first describe context encoder and query encoder in Section \ref{ssec:model:encoder}, neural topic model in Section \ref{ssec:model:topic-model}. Then the topic-aware decoder and learning objective will be in turn given in Section \ref{ssec:model:decoder} and \ref{ssec:model:objective}. 

\subsection{Interaction Modeling}\label{ssec:model:encoder}

To describe turn interactions, we first assume that there are $m$ chronologically-ordered turns given as contexts and each turn $T_i$ is formulated as a word sequence $\langle w_{i,1}, w_{i,2}, ..., w_{i,n_i}\rangle$ ($n_i$ denotes the number of words).
We consider the $m$-th turn as the query while others the history ($T_{history}=\langle T_1, T_2, ..., T_{m-1}\rangle$). 
Here we distinguish the query and its earlier history to separately explore the quote's language coherence to the query (for query consistency) and its interaction consistency to the earlier posted turns. 
In the following, we will describe how to encode history and query turns, and how the learned representations work together to explore conversation structure.  

%Formally, given a conversation with m turns. The first m-1 turns are marked as . Each turn $T_i \in T_{history}$ is in the form $\langle w_{i,1}, w_{i,2}, ..., w_{i,n_i}\rangle$, where $N_i$ is the number of words in $T_i$. The last turn is regarded as query, marked as $T_{query} = \langle w_{m,1}, w_{m,2}, ..., w_{m,n_m}\rangle$. 

\paragraph{History Encoder.}
Here we describe how to encode turns in history.
We first feed each word $w_{ij}$ (the $j$-th word in the $i$-th turn) in history into an embedding layer to obtain its word vector $\bm{c}_{i,j}$.
%, where i indicates $i^{th}$ turn and j indicates $j^{th}$ word. 
Then word vectors of the $i$-th turn $C_i = \langle \bm{c}_{i,1}, \bm{c}_{i,2}, ..., \bm{c}_{i,n_i}\rangle$ are further processed with a bidirectional gated recurrent unit (Bi-GRU) \cite{cho2014learning}. Its hidden states are defined as:
\begin{equation}\small
\overrightarrow{\bm{h}^c_{i,j}} = f_{GRU}(\bm{c}_{i,j}, \bm{h}^c_{i,j-1}), \,\overleftarrow{\bm{h}^c_{i,j}} = f_{GRU}(\bm{c}_{i,j}, \bm{h}^c_{i,j+1})
\end{equation}
%\vskip -0.2em
%\vskip -0.2em
%\begin{equation}\small
%\overleftarrow{\bm{h}^c_{i,j}} = f_{GRU}(\bm{c}_{i,j}, \bm{h}^c_{i,j+1})
%\end{equation}
%\vskip -0.2em
\noindent The turn-level representations are hence captured by concatenating the last hidden states of both directions: $\bm{h}^c_i = [\overrightarrow{\bm{h}^c_{i,n_i}}; \overleftarrow{\bm{h}^c_{i,0}}]$. 
Further, we define the history representations as $\bm{h}^c = \langle \bm{h}^c_1, \bm{h}^c_2, ..., \bm{h}^c_{m-1}\rangle$, which will be further used to encode the interaction structure (described later).

\paragraph{Query Encoder.}

Similar to the way we encode each turn in history, a Bi-GRU is first employed to learn query representations $\bm{q} = \bm{h}^c_m$.
Then, we identify which turns in history the query responds to for learning interaction consistency.
To this end, we put a query-aware attention over the history turns and result in a context vector below:
\begin{equation}
\bm{c} = \sum_{i=i}^{m-1}\alpha_i \cdot \bm{h}_i^c,\ \ \ \alpha_i = softmax(\bm{h}_i^c \cdot \bm{q})
\end{equation}
%Then to construct a more informative query that is aware of history context, we apply an attention mechanism here, where dot product is adpoted for attention weights computation:
%\vskip -0.2em
%\begin{equation}\small
%\bm{c} = \sum_{i=i}^{m-1}\alpha_i \cdot \bm{h}_i^c,\ \ \ \alpha_i = softmax(\bm{h}_i^c \cdot \bm{q})
%\end{equation}

\noindent Afterwards, we enrich query representations with the features from history and obtain the history-aware query representations:
\begin{equation}
\Tilde{\bm{q}} = W_q[\bm{q} ; \bm{c}] + b_q
\end{equation}
\noindent where $W_q$ and $b_q$ are learnable parameters.

\paragraph{Structure Encoder.}
With the representations learned above for query $\Tilde{\bm{q}}$ and history $\bm{h}^c$, we can further explore how turns interact with their neighbors (henceforth conversation structure) with another Bi-GRU. 
It is fed with the $\langle \bm{h}_1^c, \bm{h}_2^c, ..., \bm{h}_{m-1}^c, \Tilde{\bm{q}}\rangle$ sequence and the hidden states sequence $\langle\bm{h}_1, \bm{h}_2, ..., \bm{h}_{m-1}, \bm{h}_m\rangle$ is further put into a memory bank $M$ for decoder’s attentive retrieval in quotation generation (see Section~\ref{ssec:model:decoder}).

\subsection{Topic Modeling}\label{ssec:model:topic-model}
Following the common practice~\cite{Blei:2003:LDA:944919.944937,pmlr-v70-miao17a}, we model topics following the bag-of-words (BoW) assumption.
Hence, we form a BoW vector ${\bf x}_{bow}$ (over vocabulary $V$) of the words in context to learn its discussion topic.
%
%Before we process into Topic Modeling, we first construct global bag-of-words representation from the whole corpus. Specifically, given the conversation corpus $\mathcal{C}$ with $|\mathcal{C}|$ conversations, we count up each term appeared to construct ${\bf x}_{bow}$, a bog-of-words (BoW) verctor over vocabulary $V$.
%Formally, given a collection $\mathcal{C}$ with $m$ conversations, $\{{\bf x}_1, {\bf x}_2, ..., {\bf x}_{m}\}=\{T_{context}, T_{query}\}$, each dialogues $\bf x$ will be processed into the bag-of-words (BoW) form ${\bf x}_{bow}$. ${\bf x}_{bow}$ is the BoW term vector over the vocabulary $V$. 
%${\bf x}_{bow}$ (following the bag-of-words assumption in most topic models~\cite{Blei:2003:LDA:944919.944937,pmlr-v70-miao17a}) serves as the input for topic modeling. 
The topic inference process is inspired by neural topic models (NTM) \cite{pmlr-v70-miao17a}. It is based on a variational auto-encoder (VAE)~\cite{DBLP:journals/corr/KingmaW13} involving an encoding and a decoding step to reconstruct the BoW of contexts.

\paragraph{BoW Encoding Step.} 
This step is designed to learn a latent topic variable $\bf z$ from ${\bf x}_{bow}$.
Here words in conversation contexts are assumed to satisfy a Gaussian distribution prior on mean $\mu$ and standard deviation $\sigma$~\cite{pmlr-v70-miao17a}.
 They are estimated by the following formula:
\begin{equation}\
    {\bf \mu}=f_{\mu}(f_e({{\bf x}_{bow}})) , \,    log\,{\bf \sigma}=f_{\sigma}(f_e({{\bf x}_{bow}}))
\end{equation}
\noindent where $f_*(\cdot)$ is a neural perceptron performing a linear transformation activated with an ReLU function~\cite{Nair:2010:RLU:3104322.3104425}.

\paragraph{BoW Decoding Step.} 

Conditioned on the latent topic $\bf z$, we further generate words to form the BoW of each conversation ${\bf x}_{bow}$. 
Here we assume each word $w_n\in {\bf x}_{bow}$ is drawn from the conversation's topic mixture $\theta$, which is a distribution vector over the topics.
In the following, we show the generation story to decode ${\bf x}_{bow}$:
\begin{compactitem}
    \item Draw latent topic ${\bf z}\sim \mathcal{N}({\bf \mu}, {\bf \sigma}^2)$.
    \item  Topic mixture $\theta=softmax(f_{\theta}({\bf z}))$.
    \item For the $n$-th word in the conversation: 
    \begin{compactitem}
        \item Draw the word $w_n\sim softmax(f_{\phi}(\theta))$.
    \end{compactitem}
\end{compactitem}

\noindent Here $f_*(\cdot)$ is a ReLU-activated neural perceptron defined above. 
The topic mixture $\theta$ will be later applied to capture topic consistency when predicting the quotation.
%The weight matrix of $f_{\phi}(\cdot)$ (with the softmax normalization) is employed as the topic-word distributions.
%We adopt the topic mixture $\theta$ as the topic representations to guide our quotable phrase generation.

\subsection{Quotation Generation}\label{ssec:model:decoder}

To predict the quotation $\bm{y}$, we first define the probability of words in it with the following formula:
\vskip -1em
\begin{equation}
    Pr(\bm{y}|T_{history}, T_{query}) =  \prod_{i=1}^{|\bm{y}|}Pr(y_i|\bm{y}_{<i}, M, \theta) 
\end{equation}

\noindent where $\bm{y}_{<i}=\langle y_1, y_2, ..., y_{i-1}\rangle$ and $|\bm{y}|$ denotes the quotation's word number.
In prediction, the $i$-th word is generated with a likelihood $p_i=Pr(y_i|\bm{y}_{<i}, M, \theta)$, which is jointly determined by the words appearing before it ($\bm{y}_{<i}$) and the contexts features delivered by $M$ (turn interactions described in Section~\ref{ssec:model:encoder} ) and $\theta$ (the discussion topic described in Section~\ref{ssec:model:topic-model}). Below comes more details of how we follow the semantic consistency of contexts to generate quotations.

%With the memory bank $M$ in Section~\ref{ssec:model:encoder} and topic distribution $\theta$ derived from Section~\ref{ssec:model:topic-model}, we define the process of generating quotable phrase with the following probability:
%\vskip -1em
%\begin{equation}\small
%    Pr(\bm{y}|T_{history}, T_{query}) = \prod_{i=1}^{|\bm{y}|}Pr(y_i|\bm{y}_{<i}, M, \theta) 
%\end{equation}
%\vskip -0.5em
%\noindent where $\bm{y}_{<i}=\langle y_1, y_2, ..., y_{i-1}\rangle$. We denote $Pr(y_i|\bm{y}_{<i}, M, \theta)$ as $p_i$, which is a word distribution over vocabulary $V$, reflecting the possibility that a word appearing in the $i^{th}$ position of the target quotable phrase. Then we will introduce the procedure of obtaining $p_i$ as the following.

\paragraph{Query Consistency.}
To carry on query's senses, the quotation is decoded with an unidirectional GRU initialized based on the encoded query. The initialization and later recursion of decoder's hidden states are given as: 
\begin{equation}
\bm{h}_0^d  = W_0\Tilde{\bm{q}}+b_0, \,  \bm{h}_i^d = f_{GRU}(\bm{v}_i, \bm{h}_{i-1}^d)
\end{equation}
%\vskip -0.5em
% Further, we will have the $i$-th hidden states as:

% %\vskip -1em
% \begin{equation}\small
%      \bm{h}_i^d = f_{GRU}(\bm{v}_i, \bm{h}_{i-1}^d)
% \end{equation}
% %\vskip -0.5em
\noindent where $W_0$ and $b_0$ are parameters to be learned. $\bm{v}_i$ is the embedded decoder input to predict the $i$-th word in quotation.\footnote{In training, we do teacher forcing and feed the gold standard. In test, we feed the predicted left neighbor.} In decoding, word prediction is conducted sequentially with beam search. It results in a ranking list of output, where we take the top $K$ for quotation matching described later.
%, and $\bm{h}_{i-1}^d$ is the hidden state of the last timestamp.

%We initiate the GRU decoder with $\bm{h}_0^d  = W_0\Tilde{\bm{q}}+b_0$, which can ensure the semantic consistency between the generated quotable phrase and query to some extent. 

\paragraph{Topic and Interaction Consistency.}
For modeling quote consistency of discussion topics (with $\theta$) and turn interactions (with $M$), we design a turn-based attention over conversation contexts to decode the quotation.
%
%To distinguish different turns interacted with target quotable phrase, and make it aware of topics, we design an attention mechanism during predicting the $i^{th}$ word in quotable phrase. 
Its attention weights are computed in aware of the structure-encoded turn representations $\bm{h}_j$ from $M$ and topic distribution $\theta$:
%Considering the Memory bank $M$ and NTM information $\theta$ at the same time, when we predict the $i^{th}$ word in quotable phrase, we can compute attention over $M$ as the following.
\begin{equation}\label{eq:attention}
     \alpha_{ij} = \frac{exp(f_d(\bm{h}_i^d,\bm{h}_j,\theta))}{\sum_{j'=1}^{m}exp(f_d(\bm{h}_i^d, \bm{h}_{j'}, \theta))}
\end{equation}
\noindent where $f_d(\bm{h}_i^d,\bm{h}_j,\theta)$ captures the topic-aware semantic dependency the $i$-th word in quotation to the $j$-th turn in contexts and is defined as:

%relations between the $j$-th turn in the source conversation and the $i^{th}$ word in the target quotable phrase. And it can be computed according to the follow formula:
\vskip -0.5em
\begin{equation}
     f_d(\bm{h}_i^d,\bm{h}_j,\theta) = W_d[\bm{h}_i^d\cdot \bm{h}_j^{\theta}]+b_d
\end{equation}
%\vskip -0.5em
\noindent where $\bm{h}_j^{\theta}=W_{\theta}[\bm{h}_j;\theta]+d_{\theta}$, and parameters $W_d$, $b_d$, $W_{\theta}$, and $d_{\theta}$ are all trainable. 
%$f_d$ measures the semantic relations between the $j^{th}$ turn in the source dialogue and the $i^{th}$ word in the target quotable phrase. 
Then we give the context vector $\bm{t}_i$ conveying both topic and interaction features for the $i$-th word to be generated:
\vskip -0.5em
\begin{equation}
     \bm{t}_i = \sum_{j=1}^{m}\alpha_{ij}\bm{h}_j.
\end{equation}
Finally, we predict the $i$-th word in quotation following the distribution $p_i$ defined to combine topic, interaction, and query consistency:

% the distribution $p_i$ over vocabulary $V$ for predicting the $i^{th}$ target word with the formula below:
\vskip -1em
\begin{equation}
     p_i = softmax(W_{p}[\bm{h}_i^d;\bm{t}_i]+b_{p}),
\end{equation}
%\vskip -0.5em
\noindent where $W_p$ and $b_p$ are trainable parameters. 

%\lingzhi{
\paragraph{Quotation Matching.}
Occasionally language generation will ``create'' a non-existing quotation. To avoid that, we take a post-processing step for the outputs absent in our quotation list. Following previous practice \cite{liu-etal-2019-neural-based}, we select a quote from the list with the minimum edit distance (by tokens) and consider it as the final output.
%We map each generation result into the most related quotation in candidate list, using edit distance following \citet{liu-etal-2019-neural-based}. 
%In this way, we prevent our results from generating quotations not exactly exist in the list.
%}

\subsection{Learning Objective}\label{ssec:model:objective}
For the entire framework, we design its learning objective to allow joint learning of latent topics and conversation quotations:
\vskip -0.5em
\begin{equation}\label{eq:obj}
\mathcal{L} = \mathcal{L}_{NTM} + \mathcal{L}_{QGM}
\end{equation}
Here $\mathcal{L}_{NTM}$ is the objective function of neural topic model (NTM) defined as:
\vskip -1.em
\begin{equation}\label{eq:obj-NTM}
\mathcal{L}_{NTM} = D_{KL}(p({\bf z})||q({\bf z}\,|\,{\bf x}))-\mathbb{E}_{p({\bf z})}[p({\bf x}\,|\,{\bf z})]
\end{equation}
%\vskip -1.0em

\noindent where $D_{KL}(\cdot)$ is the Kullback-Leibler divergence loss and $\mathbb{E}_{*}[\cdot]$ reflects the reconstruction loss. \footnote{Because of the space limitation, we leave out the derivation details and refer the readers to \citet{pmlr-v70-miao17a}.}

As for $\mathcal{L}_{QGM}$, it is defined as the cross entropy loss over all training instances to train the quotation generation model (QGM):
\vskip -1em
\begin{equation}\label{eq:obj-QG}
\mathcal{L}_{QGM} = -\sum_{n=1}^{N}log(Pr(\bm{y}_n|C_n,\theta_n))
\end{equation}

\noindent where $N$ is the number of training instances. $C_n = \{T_{history}, T_{query}\}_n$ represents the contexts of the $n$-th conversation and $\theta_n$ is $C_n$’s topic composition induced by NTM.

\section{Experimental Setup}

\paragraph{Datasets.} For experiments, we construct two new datasets: one in Chinese from Weibo (a popular microblog platform in China and henceforth \textbf{Weibo}) and the other in English from Reddit (henceforth \textbf{Reddit})\footnote{The datasets are available at \url{https://github.com/Lingzhi-WANG/Datasets-for-Quotation-Recommendation}}. 
Here the raw Weibo data is released by \citet{wang-etal-2019-topic-aware} and Reddit obtained from a publicly available corpus.\footnote{\url{https://files.pushshift.io/reddit/comments/}} 
%and use the data from July 2013 to May 2015.
For both Weibo and Reddit, we follow the common practice form conversations with posts and their comments~\cite{DBLP:conf/emnlp/LiGWPW15,DBLP:conf/naacl/ZengLWBSW18}, where a post or comment is considered as a conversation turn.
%The posts and comments in Weibo or Reddit can be regarded as a kind of online conversation. 

To gather conversations with quotations, we maintain a quotation list and remove conversations containing no quotation from the list.
For the remaining, if a conversation has multiple quotes, we construct multiple instances where one corresponds to the prediction of a quotation therein.
On Weibo, we explore the quoting of Chinese \textit{Chengyu}.\footnote{\url{https://en.wikipedia.org/wiki/Chengyu} Chengyu can be seen as a quotable phrase~\cite{wang2013study} --- memorable rhetorical figures to convey wit and striking statement~\cite{DBLP:conf/clfl/BenderskyS12}.}
% context~\cite{DBLP:conf/clfl/BenderskyS12}} --- four-character phrases from ancient Chinese literature, where a list is gathered from open source Chengyu dictionary.\footnote{\url{https://github.com/pwxcoo/chinese-xinhua}} 
% to be our quotations because of their essential nature to be quotable~\cite{wang2013study} --- as rhetorical figures of wisdom patterns to convey persuasive and memorable statement able to be quoted without its original
% context~\cite{DBLP:conf/clfl/BenderskyS12}. 
For Reddit, we obtain the quotation list from Wikiquote.\footnote{\url{https://en.wikiquote.org/wiki/Main_Page}}
Afterwards, we remove conversation instances with quotations appearing less than $5$ times to avoid sparsity~\cite{tan2015learning}. Finally, the datasets are randomly splitted into $80$\%, $10$\%, and $10$\%, for training, development, and test.

\begin{table}[t]
\newcommand{\tabincell}[2]{\begin{tabular}{@{}#1@{}}#2\end{tabular}}
\begin{center}
\begin{tabular}{|l|rr|}
\hline 
&  \textbf{Weibo} & \textbf{Reddit} \\
\hline

\hline
{\# of quotes}& 1,053 & 1,111  \\
{Avg len of quotes}& 4.0 & 10.1  \\%max 63
{$|$Voc$|$ of quotes}& 1,251 &  4,111 \\
\hline
\hline
{\# of convs} & 19,081 & 44,539  \\
{Avg \# of turns per conv}& 2.51 & 4.25  \\
{Avg len of turn per conv}&21.6 &71.8 \\
%{$|$Voc$|$ of convs}& 43822 & 70414  \\
{$|$Voc$|$ of convs}& 44,134 & 72,375  \\
\hline
\end{tabular}
\end{center}
 \caption{
 \label{statistics-table1}Statistics of Weibo and Reddit datasets. The upper rows are for quotes and the lower rows are for conversations.
 The ``len'' refers to the number of tokens contained. ``Avg \# of turns'' means the average turn number in context.
% Avg len: average number of tokens. Avg \# of turns per conv: average number of turns in a conversation. 
%$|$Voc$|$: the vocabulary size. 
}
\end{table}
%To build the datasets, we firstly collect idioms, English quotes. 
%For Weibo, we constructed conversations from tweets dataset, 	
%Tweets2011\footnote{\url{https://trec.nist.gov/data/tweets/}}\shan{weird...}, and conversations contain idioms. 

\begin{figure}[ht]
\centering
\subfigure[Frequency Distribution]{\label{sfig:data:fre}
\includegraphics[width=3.6cm]{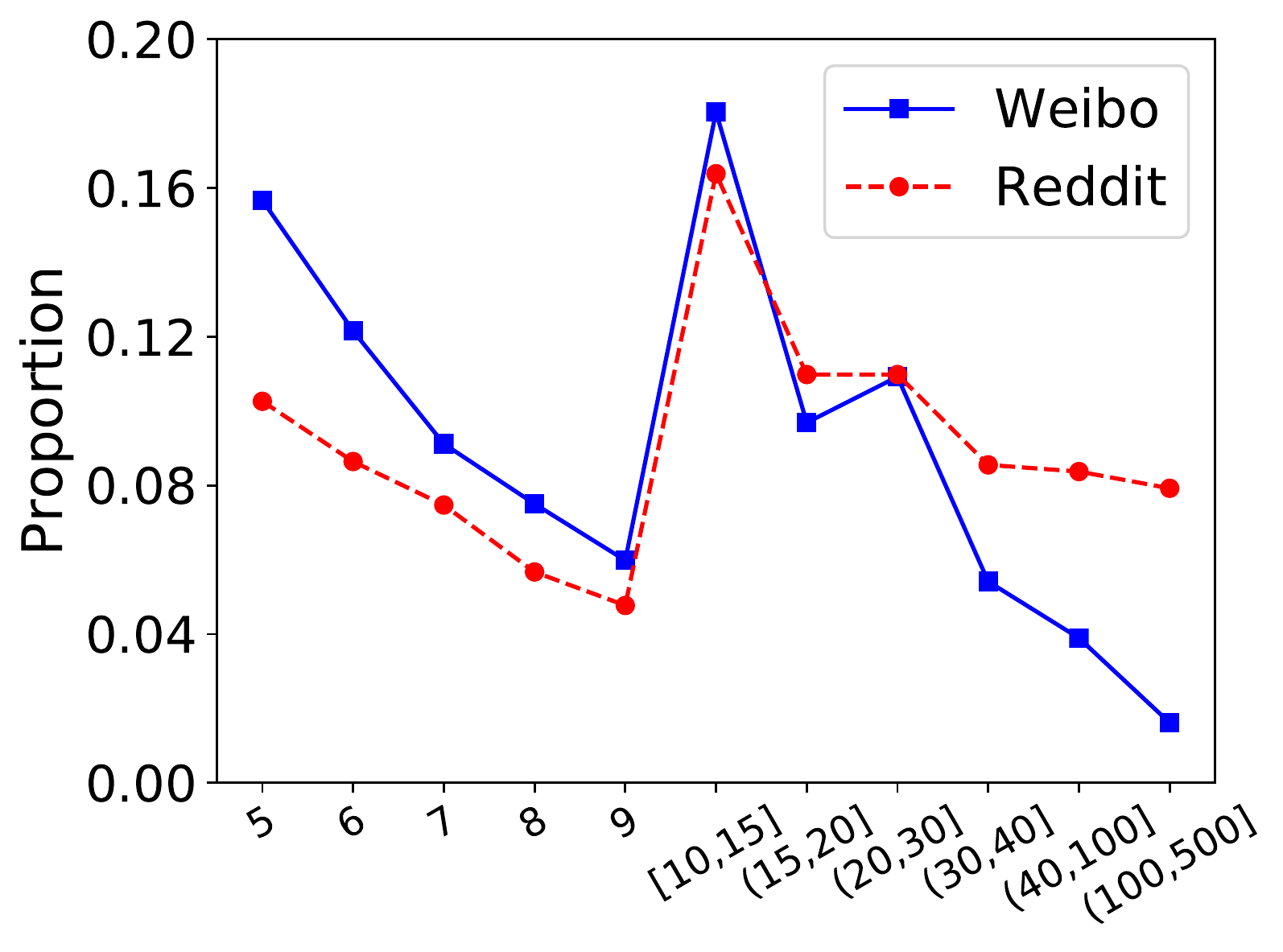}
}
\subfigure[Position Distribution]{\label{sfig:data:turn}
\includegraphics[width=3.6cm]{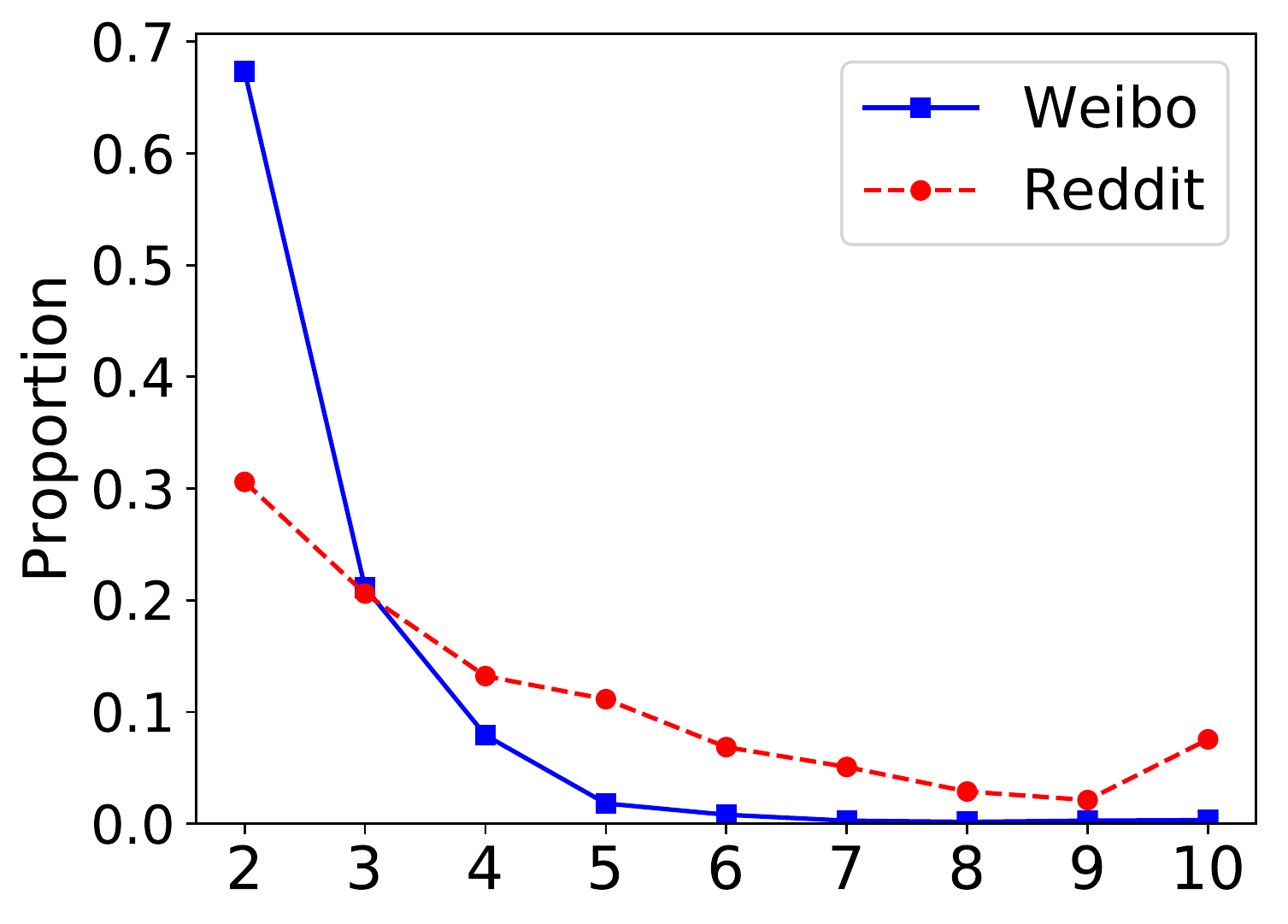}
}
\caption{
\label{fig:data:distribution}
Quotation distribution over frequency (on the left) and position (right). X-axis: frequency (left) and context turn number (right); Y-axis: proportion of quotations (left) and conversations (right).
%Left: the distribution on the frequency of quotations, where x-axis indicates the frequency of quotations and y-axis indicates the proportion.
%Right: the distribution of turn number in conversation contexts, where x-axis indicates turn number y-axis proportion. Reddit exhibits flatter distribution.
}
\end{figure}

%We regard every post or comment as a single turn of conversation and use the turns in front of quotations and quotations themselves as our data. 
%The minimum number of turns of Weibo and Reddit are both 2, and maximum number are both 10. 
%After filtering quotations that appear
%less than 5 times, there are 1053 and 1111 unique quotes for Weibo and Reddit respectively. 
The statistics of the two datasets are shown in Table \ref{statistics-table1}. 
%As can be seen, the conversation number is much larger than quote. It indicates that only a few quotations will be commonly used in online conversations, probably because of its informal writing style.
%We can see that the number of quotations we use in real on-line conversations is limited, but the correct use of these limited quotations can add a lot of luster to our conversations. 
We observe that the two datasets exhibit different statistics.
For example, from the average turn number in contexts, we find Reddit users tend to quote in later turns while Weibo earlier.
To further compare users' quoting behavior, we show the distribution of quotation number in Figure \ref{sfig:data:fre} and position in Figure \ref{sfig:data:turn}. 
%displays the turn number distribution in contexts 
Figure \ref{sfig:data:fre} shows only a few quotations are commonly used in online conversations, probably because of its informal writing style.  
While for Figure \ref{sfig:data:turn}, we find only a few Weibo conversations quote $5$ turns later while the distribution on Reddit is much flatter. %Figure \ref{sfig:data:fre} shows the distribution of frequency. We can see that most of the quotations' frequency is less than 40.

% and we can find that the data with more than 5 turns is very sparse in Weibo dataset. Instead, the data on reddit is more evenly distributed.

%We have also tried to ensure that each quotation will appear in the training, development and test set in the proportion of 8:1:1, and found that the performances of models are similar to that of not ensuring that each  appears in each set. 

% \begin{figure}[t]
% \centering
% \includegraphics[width=0.45\textwidth]{pictures/data_turn.pdf}
% \vskip -0.5em
% \caption{\label{fig:data_turn} The distribution of turn number in conversation contexts, where x-axis indicates turn number y-axis proportion. Reddit exhibits flatter distribution. }
% \vskip -1em
% \end{figure}
\paragraph{Preprocessing.} 
To preprocess Weibo data, we adopted open-source Jieba toolkit\footnote{\url{https://github.com/fxsjy/jieba}} for Chinese word segmentation. 
For Reddit dataset, we employ natural language toolkit (NLTK\footnote{\url{https://www.nltk.org}}) for tokenization. 
In BoW preparation, all stop words and punctuation were removed following common practice to train topic models~\cite{ Blei:2003:LDA:944919.944937}.

\paragraph{Parameter Setting.} 
%The QGM model is set up mostly based on \cite{wang-etal-2019-topic-aware}. 
Here we describe how we set our model. 
In model architecture, the hidden size of all GRUs is set to $300$ (bi-direction, $150$ for each direction). 
For encoder, we adopt two layers of bidirectional GRU, and unidirectional GRU for decoder. 
The parameters in NTM are set up following \citet{zeng2018topic}. 
For input, we set the maximum turn length to $150$ for Reddit and $200$ for Weibo, and the maximum quotation length $20$.
Word embeddings are randomly initialized to $150$-dimensional vectors.
\begin{table*}[t]\setlength{\tabcolsep}{2.5mm}
\newcommand{\tabincell}[2]{\begin{tabular}{@{}#1@{}}#2\end{tabular}}
\begin{center}
\scalebox{0.9}{
\begin{tabular}{|l|rrrrr|rrrrr|}
\hline 
\multirow{2}{*}{Models} 
&\multicolumn{5}{c|}{ \tabincell{c}{\textbf{Weibo}} }
&\multicolumn{5}{|c|}{ \tabincell{c}{\textbf{Reddit}} }

\\
\cline{2-11}
& P@1 & MAP & RG-1 & RG-L &BLEU& P@1 & MAP & RG-1 & RG-L &BLEU\\
\hline
\hline
\underline{\textbf{Weak Baselines}} & & &&& & & &&&\\
\textsc{Random} &0.2 &0.7&2.1&2.1&0.2&
0.1&0.7&5.6&4.5&0.1\\
\textsc{Frequency} & 2.3 & 6.9 & 3.1 & 3.1&2.3
& 1.0 & 4.7 &1.7 &1.5&1.0\\
\hline
\hline
\underline{\textbf{Ranking Models}} & & &&& & & &&&\\
\textsc{LTR} & 3.6 & 9.3 & 5.1& 5.1&3.6
 & 1.7 & 7.1 & 4.1  & 3.6&1.7\\
\textsc{CNN-LSTM} & 7.3&11.3&10.5&10.5&7.3
&4.1&5.2&6.8&6.0&3.7\\
% \textsc{BERT-LSTM} &8.9 &14.4 &10.8 &10.9 &9.3
% &14.3 &22.4  &19.1& 19.2 &11.7\\
\hline
\hline
\underline{\textbf{Generation Models}} & & &&& & & &&&\\
\textsc{Seq2Seq}&19.9&24.1&22.6&22.5&19.9
&7.2&9.8&11.7&10.6&4.7\\
\textsc{TAKG}&24.0&27.3&26.8&26.7&24.0
&12.5&16.0&15.7&14.4&6.7\\
\textsc{NCIR} &22.6 &26.5 &25.3 &25.2&22.6 
&7.3 & 12.2 & 10.9 & 9.9&4.1\\
\hline
\hline
\underline{\textbf{Our models}}&&&&&&&&&&\\
\textsc{IE only}  &21.5 & 24.8  & 24.5 &24.4&21.5
&11.2 &14.6 &13.9 &12.8 &5.7\\
\textsc{IE+QE}  &22.0 &24.7 &25.2 &25.1&22.0
&13.5 &17.4 &17.0 &15.5&7.0\\
\textsc{IE+QE+NTM} &\textbf{27.2}$^\ddagger$ & \textbf{31.6}$^\dagger$ & \textbf{29.5} $^\ddagger$&\textbf{29.5}$^\ddagger$&\textbf{27.2}$^\ddagger$
& \textbf{17.5}$^\dagger$& \textbf{24.0}$^\dagger$ & \textbf{20.3}$^\dagger$
&\textbf{18.8}$^\dagger$&\textbf{9.5}$^\dagger$\\
\hline
\end{tabular}
}
\end{center}
\caption{\label{tab:main} Comparison results on Weibo and Reddit datasets (in \%). RG-1 and RG-L refer to ROUGE-1 and ROUGE-L respectively. The best results in each column are in \textbf{bold}. Our full model \textsc{IE+QE+NTM} achieves significantly better performance than all the comparisons (paired t-test. $\ddagger$: $p<0.05$; $\dagger$: $p<0.01$)
}
\end{table*}
In model training, we employ Adam optimizer ~\cite{kingma:adam}, with $1e-3$ learning rate and the adoption of early stop~\cite{caruana2001overfitting}. 
Dropout strategy~\cite{Srivastava:2014:DSW:2627435.2670313} is also used to avoid overfitting. 
%For quotation generation,
We adopt beam search (beam size $=5$) to generate a ranking list for quote recommendation.

\paragraph{Evaluation Metrics.}
We first adopt recommendation metrics with popular information retrieval metrics Precision at K (P@K) and mean average precision (MAP) scores ~\cite{schutze2008introduction} used. % following \citet{wang-etal-2019-topic-aware}. 
For P@K, K=1 to measure the top prediction, while for MAP we consider the top $5$ outputs. Here we measure the generation models with their predictions after quotation matching (Section \ref{ssec:model:decoder}).
%\lingzhi{
%For generation models, the P@1 and MAP scores are based on the post-processed results (see Section \ref{ssec:model:decoder}).
%} 
%Since different K values are tested on P@K and result in a similar trend, so only P@1 are reported, which is the most import in recommendation task. 
%MAP scores are computed given the top 5 outputs. 
%For Reddit dataset, since the maximum length of generated quotable phrases is set to 20, we think the generated sentence is correct as long as the generated quotable phrase is a substring of ground truth. 
Then, generation metrics are employed to evaluate word-level predictions. Here we consider both ROUGE~\cite{lin-2004-rouge} from summarization (F1 scores of ROUGE-1 and ROUGE-L are adopted) and BLEU~\cite{papineni-etal-2002-bleu} from translation. 
%For ROUGE, we report the F1 scores of ROUGE-1 and ROUGE-L.
To allow comparable results, generation models are measured with their original outputs (without quotation matching) while for ranking competitors, we take their top-1 ranked quotes. 
%\lingzhi{For generation models, we leverage the generated quotations  before post-processing to compute BLEU and ROUGE. For ranking models, we use the top-1 ranked quotations to compute BLEU and ROUGE scores.}

%Besides, metrics for summarization or generation evaluation are also adopted, where ROUGE~\cite{lin-2004-rouge} (ROUGE-1 and ROUGE-L) and BLEU~\cite{papineni-etal-2002-bleu} are reported. 
%After experiment we find BLEU~\cite{papineni-etal-2002-bleu} shows the similar trending with ROUGE, we did not list the experimental results to save space.

\paragraph{Comparisons.} 
%For comparisons, we first consider a weak baseline 1) RANDOM that randomly rank quotations from training data. Then We adopt a baseline 2) FREQUENCY that rank quotations according to their frequency in the training set. Two models that are not seq2seq-based are also considered, learning to rank(henceforth LTR)~\cite{tan2015learning} and CNN-LSTM~\cite{Lee:2016:QRD:2911451.2914734}. For LTR, we can only use 10 of the 16 features mentioned in the \cite{tan2015learning} to do experiment. We compare with a Seq2Seq model \cite{meng2017deep} and topic-aware generation model \cite{wang-etal-2019-topic-aware} in a different task, keyphrase generation. We also compare with the state-of-the-art model \cite{liu-etal-2019-neural-based} in a related task, idiom recommendation. 

We first adopt two weak baselines that select quotations unaware of the target conversation: 1) \textsc{\underline{Random}}: selecting quotations randomly;
2) \textsc{\underline{Frequency}}: ranking quotations with frequency. 
Then, we compared two ranking baselines: 
3) non-neural learning to rank model (henceforth \textsc{\underline{LTR}}) with handcrafted features proposed in ~\citet{tan2015learning}.
%and we adopted 10 of the 16 features mentioned in the \cite{tan2015learning} to do experiment. 
4) \textsc{\underline{CNN-LSTM}}~\cite{Lee:2016:QRD:2911451.2914734}: previous quotation recommendation model (CNN for turn and quotation encoding and LSTM for conversation structure).
% \lingzhi{5) \textsc{\underline{BERT-LSTM}}: We treat the task as a multi-classification task and adopt BERT~\cite{devlin2018bert} to encode each turn's text, followed by an LSTM layer modeling the conversation structure.}
%modeling the conversations with a sequential model, based on the combination of CNN and RNN.
%a deep learning method which combines recurrent neural network and convolutional neural network and constructs a sequence model for the dialog thread.

Next, we consider the encoder-decoder generation models without modeling conversation structure: 
5) \textsc{\underline{Seq2Seq}}~\cite{DBLP:journals/corr/ChoMGBSB14}: using an RNN for encoding and another RNN for decoding;
%consisting of one RNN encoder for representation learning and one RNN decoder to generate the recommended quotes. 
6) \textsc{\underline{TAKG}}: Seq2Seq framework incorporating latent topics for decoding.
%a topic aware generation model used in social media keyphrase generation task. 
7) the state-of-the-art (SOTA) model \textsc{\underline{NCIR}} \cite{liu-etal-2019-neural-based} designed for Chinese idiom generation.
%using translation techniques for idiom recommendation, where idioms are treated as one pseudo target language.
%where consider idioms are written in one pseudo target language.

Finally, the following of our variants are test:
8) \textsc{\underline{IE only}}: using interaction modeling results for decoding (w/o topic and query consistency modeling);
%different from \textsc{Seq2Seq} \cite{DBLP:journals/corr/ChoMGBSB14}, 
%only encodes interactions with turn encoder and turn-based attention, regarding interaction consistency. 
9) \textsc{\underline{IE+QE}}: coupling interaction and query consistency (w/o NTM used for topic consistency);
%only considering interaction and query consistency, where query encoder is added to \textsc{IE only}. 
10) \textsc{\underline{IE+QE+NTM}}: our full model. 
%our full model where topic, interaction and query consistency are all considered, with Neural Topic Model added to \textsc{IE+QE}.

\section{Experimental Results}

In this section, we first show the main comparison results in Section~\ref{ssec:res:comparison}.
Then Section~\ref{ssec:res:interpretation} discusses what we learn to represent %topic, interaction, and query 
consistency. % and how they affect quotings.
Finally, Section~\ref{ssec:res:discussion} presents more analysis to characterize quotations in online conversations.
%examined for detailed effects exploration.

\subsection{Main Comparison Results}\label{ssec:res:comparison}

Table \ref{tab:main} reports the main comparison results on two datasets, where our full model significantly outperforms all comparisons by a large margin. Several interesting observations can be drawn:

$\bullet$~\textit{Quotation is related with context.}
The poor performance of weak baselines reveals the challenging nature of quoting in online conversations. It is not possible to learn what to quote without considering context.
%We observe that all methods exhibit less than $35$ in all metrics, especially for those weak baselines with simple strategy, indicating it is a challenging task. 
%We can observe that simple baselines such as \textsc{Random} and \textsc{Frequency} give poor performance. This indicate that quotation recommendation is a challenging task. It is difficult to rely on simple strategy to solving this task very well.

%$\bullet$~\textit{Seq2Seq-based models outperform other counterparts.}
%Compared to classification models such as \textsc{LTR} and \textsc{CNN-LSTM}, Seq2Seq-based generation models perform much better. The reasons can be described as follows. For feature-based model (\textsc{LTR}), the classification is not comprehensive and effective due to the artificial feature extraction. For \textsc{CNN-LSTM}, though it models the context by simple neural network, it can't make use of the information in the quotation. Instead, Seq2Seq-based models can utilize the information of context and quotation at the same time to some extent. 
$\bullet$~\textit{Generation models outperform Ranking.}
Generation models in encoder-decoder style perform much better than ranking. 
It maybe attributed to generation model's ability to learn word-level mapping from source context to quotation. 
%both context and quotation information, 
%while Ranking-based models depend on inefficient hand-crafted features (\textsc{LTR}) or simple structure networks (\textsc{CNN-LSTM}).

% $\bullet$~\textit{Turn-based attention, query encoder and topic model are helpful.}
% As shown in Table \ref{tab:main}, compared to \textsc{Seq2Seq}'s result \textsc{TB only} performs better, which means that modeling the conversation turn by turn is better than word by word. This is maybe due to the large granularity of information in online conversation. Compared to \textsc{TB only}'s result, \textsc{TB+QE} performs better. It can be inferred from that quotation and query are most semantically related and have some consistency. Compared to \textsc{TB+QE}'s result \textsc{TB+QE+TOPIC} performs better, which shows the importance of jointly training with NTM. This may be related to that topic information can help turn's modeling and also keep the topic of quotation consistent with that of conversation. 
$\bullet$~\textit{Interaction, query, and topic consistency are all useful.}
We see \textsc{IE only} outperforms \textsc{Seq2Seq}, showing that interaction modeling helps encode indicative features from context. 
Likewise, the results of \textsc{IE+QE} are better than \textsc{IE only}, and \textsc{IE+QE+NTM} better than \textsc{IE+QE}, both suggesting that learning query and topic consistency contribute to yield a better quotation.

$\bullet$~\textit{Quoting in Reddit is more challenging than Weibo Chengyu.}
All models perform worse on Reddit than Weibo. The possible reason is that Chinese Chengyu is shorter and renders a smaller vocabulary than English quotes (see Table \ref{statistics-table1}).  
%The main reason is that generating Chinese idioms is easier than English quotes, since Chinese idioms' lengths are mostly 4, while English quotes are longer and diverse, ranging from 4 to 63. 
%Besides, vocabulary size is also larger on Reddit. 
%In such situation, the improvement achieved by our model over other comparisons is palpable on Reddit. This may attribute to our turn-based interaction modeling, which can better handle Reddit dataset with longer average turn length.

% We can observe. that the performance on Reddit is poorer than Weibo. Besides the differences between platforms, the main reason is that the generation of Chinese idiom is easier than that of English quote. Chinese idiom's length is mostly 4, while English quote's length is much longer, ranging from 4 to 63. And English quote's vocabulary size is larger than Weibo's. 

% $\bullet$~\textit{The performance improvement of our model on Reddit dataset is more significant.}
% Though quotation recommendation is more challenging for Reddit, Our model's performance improvement compared with baselines is better in Reddit than Weibo. It may relate to the average turn length of Reddit is larger than Weibo, and our model's turn modelling can perform better in such datasets. More discussion on the impacts of turn length to our model can be found in section \ref{ssec:res:interpretation}.

\subsection{Quotation and Consistency}\label{ssec:res:interpretation}
We have shown our effectiveness in main results. Here we further examine our learned consistency and their effects on quoting. 
In the rest of this paper, without otherwise specified, \textit{our model} is used as a short form of our full model (\textsc{IE+QE+NTM}).
For comparison, we select \textsc{TAKG} for its best performance in Table \ref{tab:main} over all comparison models.

%different modules in our model, interpreting the effects of three consistencies.
\begin{figure}[t]
\centering
\includegraphics[width=0.45\textwidth]{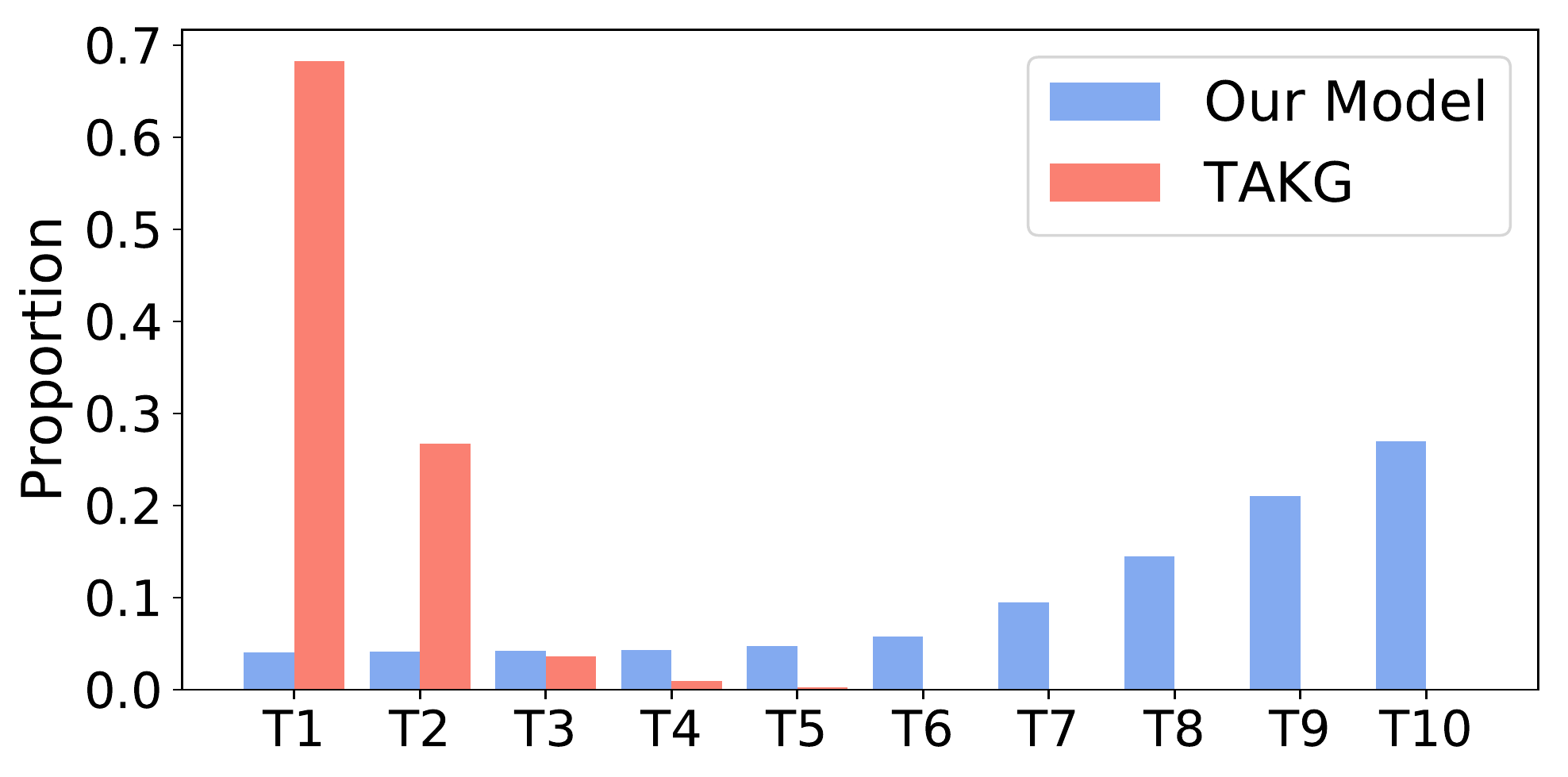}
\caption{\label{fig:attention:reddit} Attention weights over turns. X-axis: turn position; Y-axis: the normalized weight. 
%``Tn'' in X-axis means the $n^{th}$ turn of the conversation.
} 
\end{figure}

\paragraph{Interaction Consistency.}

%To capture the relative turns in interaction modeling, our model adopts turn-based attention over conversation context. 
To understand the positions of turns a quote is likely to respond to, we display the turn-based attention weights (Eq. \ref{eq:attention}) over turn position in Figure~\ref{fig:attention:reddit}.
Also shown is the attention weights from \textsc{TAKG}~\cite{wang-etal-2019-topic-aware} for comparison. 
Here we use Reddit conversations for interpretation because they involve larger turn number (see Table \ref{statistics-table1}).  
It is seen that \textsc{TAKG} can only attend the first three turns while we assign higher weights to turns closer to query.
In doing so, the quotes will continue senses from later history, which fits our intuition that participants tend to interact with latest information.

%to show whether our model indeed capture important interaction turns, together with results from the best baseline \textsc{TAKG}, which applies word-based attention and we aggregate the weights of words in same turn. 
%The conversations in Weibo mostly contain few turns, resulting in turns closed to original posts more important. Both \textsc{TAKG} and our model capture such tendency, while our model focuses more on original posts. 
%On Reddit, conversations are longer, and turns closed to the query turn will relate more in content consistency. Our model exactly holds such weights, superior than \textsc{TAKG}, benefited from turn-based interaction consistency.
%More experiments on turn lengths are described in section \ref{ssec:res:discussion}.
%Our model adopts a turn-based attention to modeling the conversation. Figure \ref{fig:turn_attention} shows the attention distribution on turns conducted on two datasets by our model and the best baseline TAKG. We can observe that our model's attention is more focused on the later parts of the dialogue, which is in the line with the closer information to quotation is more important. The conversations in Weibo are mostly short turn dialogues, which results in the attention showing the characteristics of big head and small tail. For Reddit dataset, since the length of turn is longer, the attention distribution obviously indicates the turn-based attention mechanism's advantage. More interesting experiment on the length of turns can be find in section \ref{ssec:res:discussion}. 

\paragraph{Query Consistency.}
%Table \ref{tab:main} shows that \textsc{IE+QE} performs better than \textsc{IE only}, 
We carry out a human evaluation to test the coherence of query and the predicted quotations. 
%whether query encoder can help maintain the language coherence, 
%to some extent because the decoder GRU's initiate state is passed by query encoder. 
%Because there is no mature metrics for calculating coherence at present, 
%we conduct a human evaluation to test the coherence difference between \textsc{IE+QE} and \textsc{IE only}. 
$100$ conversations are sampled from Weibo and two native Chinese speakers are invited to examine whether a quote carry on the query's senses (``yes'') or not (``no''). 
Table \ref{tab:human_eval} shows the count of ``yes'' for the ground truth quote and the output of \textsc{IE only} and \textsc{IE+QE}. 
Interestingly, even ground truth quotations cannot attain over $85$\% ``yes'', probably because of the prominent misuse of quotations on social media. 
Nevertheless, the better performance of \textsc{IE+QE} compared with \textsc{IE only} shows the usefulness to model query consistency for ensuring quotation's language coherence to the query. 
%of \textsc{IE+QE} and \textsc{IE only} 
%with their query turns on Weibo test set. 
%Then, we invite two humans who are native speakers of Chinese, to judge whether the generated quotation is coherent with query or not. 
%label '1' indicates coherence and label '0' indicates incoherence. 
%We count up the number of coherent results for both models, as well as the ground truth in Table~\ref{tab:human_eval}.
%(the number of label '1' over 100). 
%We also evaluated the results of coherence rate on the ground turn. We take 100 samples from the ground truth, \textsc{IE only}, \textsc{IE+QE} respectively, and then mix them up to let the reviewers evaluate. Table \ref{tab:human_eval} shows the result, 
%It is showed that module QE can indeed improve the coherence between quotation and query turn, reflecting the effects of query consistency for recommendation performance. 

\begin{table}[t]
\newcommand{\tabincell}[2]{\begin{tabular}{@{}#1@{}}#2\end{tabular}}
\begin{center}
\begin{tabular}{|l|c|c|}
\hline 
&  \textbf{Human 1} & \textbf{Human 2} \\
\hline
{Ground Truth}& 84 & 78 \\
\hline
\textsc{IE only}& 36 & 32  \\
\hline
\textsc{IE+QE}&49  & 46 \\

\hline
\end{tabular}
\end{center}
\caption{\label{tab:human_eval}
%Coherence results of human evaluation.
Human evaluation results for the quote coherent with query (count out of $100$).
}
\end{table}

\paragraph{Topic Consistency.}
%We employ a NTM module to capture the latent topics of conversations and make the decoder aware of topic consistency. Table~\ref{tab:main} shows that such module improves the performance (indicated by better results of \textsc{IE+QE+NTM} compared to \textsc{IE+QE}), and 
Here we use the example in Figure~\ref{fig:intro-example} to analyze the topics we learn for modeling consistency. Recall that the conversation centers around \textit{price and value} and the quote is used to argue that \textit{only fools will waste the money}. We look into the top $3$ latent topics (by topic mixture $\theta$) and display their top $10$ words (by likelihood) in Table~\ref{tab:topic}. There appears words like ``pay'' and ``stupied'', which might help to correctly predict ``fool'' and ``money'' in the quote.
\begin{table}[t]
\newcommand{\tabincell}[2]{\begin{tabular}{@{}#1@{}}#2\end{tabular}}
\begin{center}
\begin{tabular}{|c|p{5.4cm}|}

\hline

{Topic 1}& {\textcolor{blue}{\textit{game}} property child rights \textcolor{blue}{\textit{pay}} guy state church guys \textcolor{blue}{\textit{paid}}}  \\
\hline
{Topic 2}& {f**k evidence sh*t guys \textcolor{blue}{\textit{stupid}} edit nice proof dude \textcolor{blue}{\textit{dumb}}}  \\
\hline

{Topic 3}& {car \textcolor{blue}{\textit{buy}} cops police scrubs gun technology shot crime energy } \\
\hline
%{Topic 4} & {things care understand questions conversation change prove evil answer practice}  \\
%\hline
%{Topic 5} & {kill weapons war soldiers guns gun civilians isis military fight }  \\
%\hline
\end{tabular}
\end{center}
\caption{\label{tab:topic}
The top $10$ words of the $3$ latent topics related to the conversation in Figure~\ref{fig:intro-example}.  Words suggesting conversation's focus are in blue and italic.
}
\end{table}
%We select the top 3 related latent topics based on topic distribution over the conversation example, each represented by top 10 topicwords in Table~\ref{tab:topic}. It can be observed that topic words are centered on same certain topics, which all related to conversation example, reflected by words in blue and italic.
%Such result shows that topic information indeed learned by the module, and is introduced as topic consistency for the model.
% Table \ref{tab:main} shows us that \textsc{IE+QE+NTM} has the best performance, and compared to \textsc{IE+QE} we can get that topic model works. Here we shows our topic model can really learn meaningful topics. We select 5 most related latent topics out of 50 topics and write down top 10 words for each latent topic. For case described in Figure \ref{fig:intro-example}, Table \ref{tab:topic} shows the topic terms for 5 latent topics. First, we can observe that topic terms in each latent topic shown in Table \ref{tab:topic} are centre on certain topics. Second, as shown in Table \ref{tab:topic}, topic terms related to conversation listed in Figure \ref{fig:intro-example} are in blue and italic. These words show that the topic represent of the conversation makes sense, thus can help to model the conversation.

\subsection{Sensitivity to Context and Quotations}\label{ssec:res:discussion}
%In this section, we discuss some factors that may affect the performance.
%We further discuss our model with a case study and some factors that affect the effect of the model. We also analyze some interesting sentences generated.
In this section, we study how varying context and quotations affect our performance.

\paragraph{The Effects of Context.}
Here we examine whether longer context will result in better results. In the following, we measure context length in terms of turn number and token number.
\begin{figure}[t]
\centering
\includegraphics[width=0.42\textwidth]{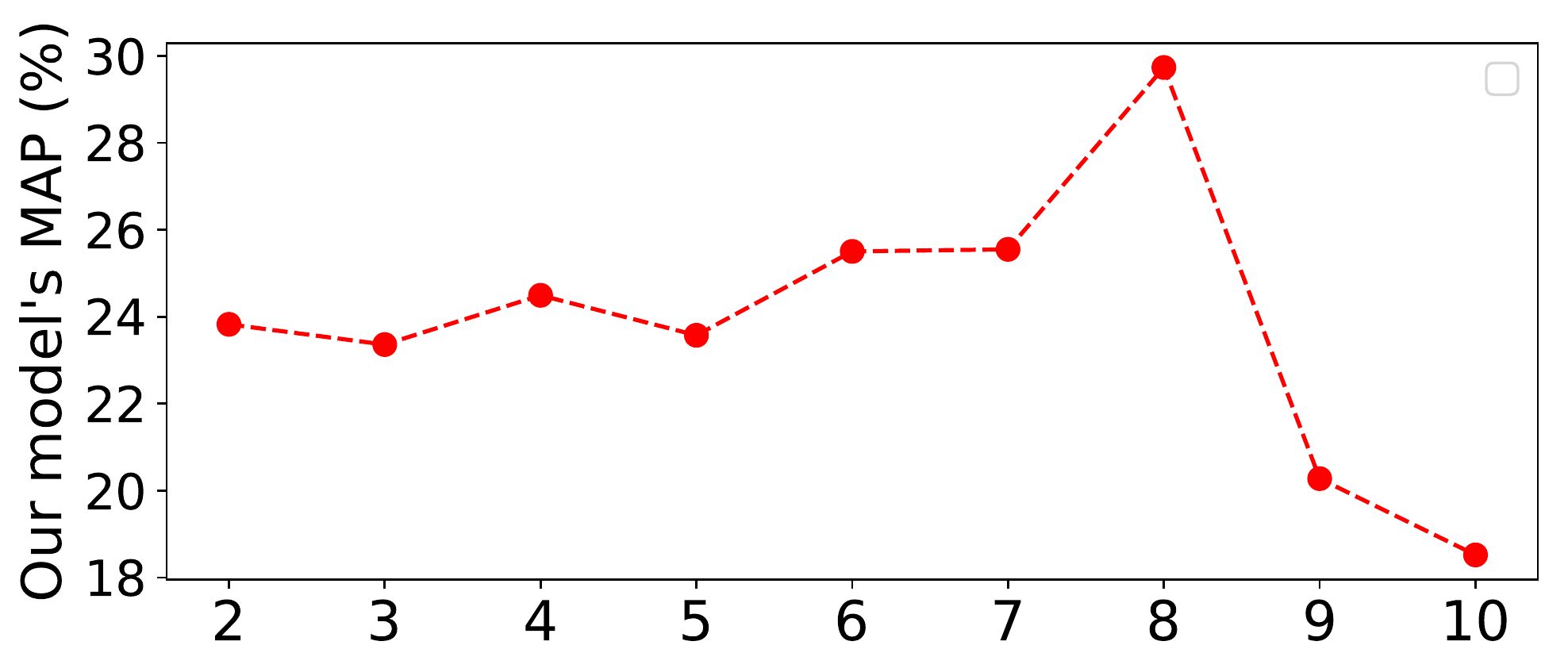}
\caption{\label{fig:turn_map} Our MAP scores over conversations with varying turn number. X-axis: turn number; Y-axis: MAP scores. The best results are seen for $8$-turn convs.
%Our model's performance on different number of turns. The size of the circle represents the amount of data.
}
\end{figure}
%Here we examine whether the length of context will affect model performance. We consider both turn number and token number.

\textit{Turn Number.}
%We compute our performance on two datasets with different number of turns, which is shown 
Figure \ref{fig:turn_map} shows our MAP scores to quote for Reddit conversations with varying turn number. 
Weibo results are not shown here for the limited data with turn number $>4$. Generally, more turns result in better MAP, for the richer information to be captured from turn interactions.
The scores drop for turn number $>8$, probably because of underfitting and a more complex model might be needed for interaction modeling.

%For Weibo dataset, the results of different turn numbers fluctuate greatly, especially for those turn numbers greater than $5$. This is due to limitation of data size, which can be reflected by circle size in Figure \ref{fig:turn_map}.
%since the amount of data is too small, the recommendation results of the dialogues whose number of turns greater than 5 fluctuate greatly, and there is no rule to follow. 
%While for Reddit dataset, our model performs generally better as turn number increases, before turn number is $8$. This reflects that, with longer context, our model can capture more information before query, and lead to better performance.
%By analysing results on Reddit Dataset, whose number distribution of data with different turn length is more average, we can find that before 8 turns, the performance of recommendation can be regarded as rising. This reflects our model performs better in longer dialogues. 

\begin{figure}[t]
\centering
\subfigure[Turn Number]{\label{sfig:turn}
\includegraphics[width=3.6cm]{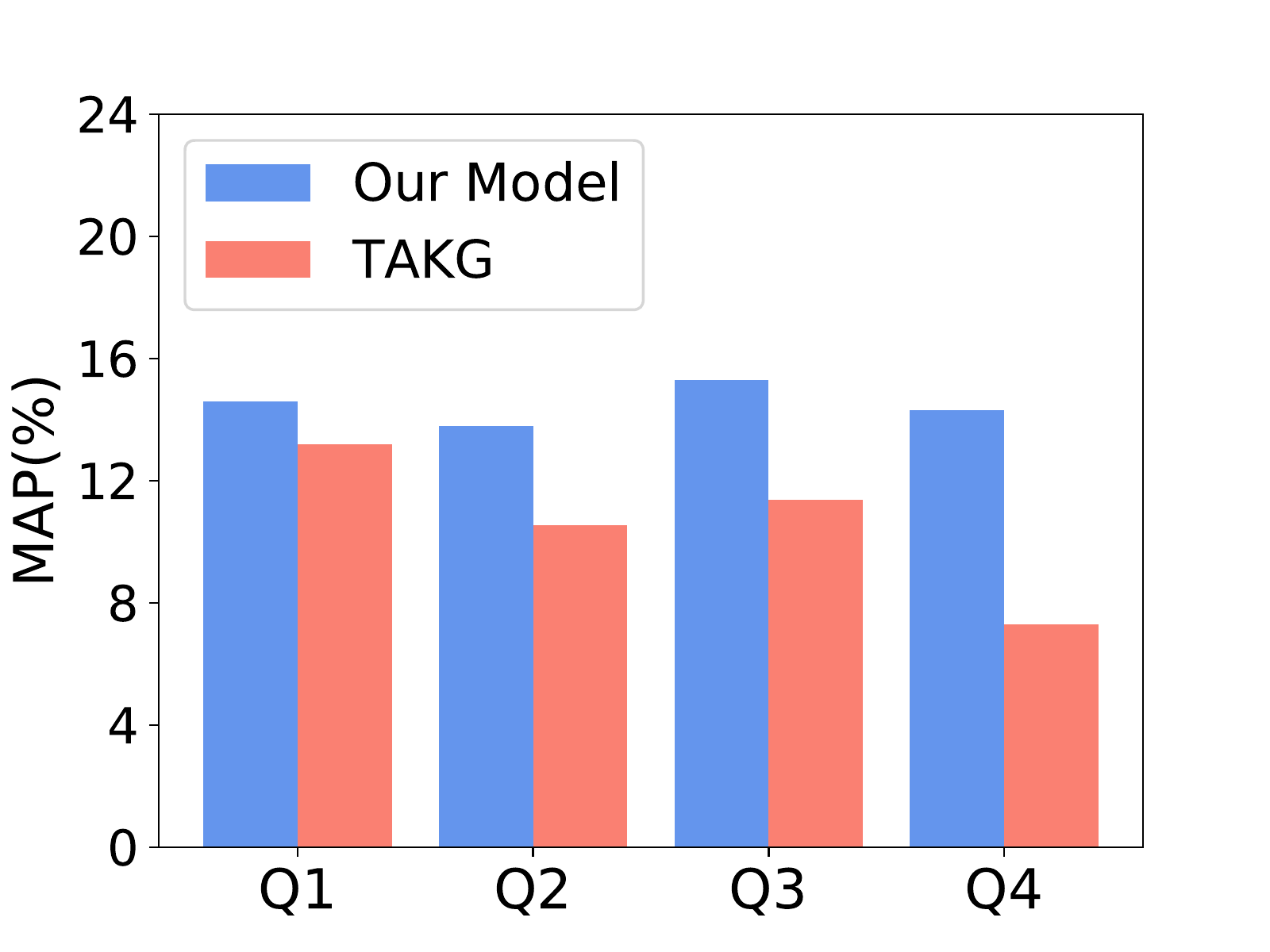}
}
\subfigure[Token Number]{\label{sfig:len}
\includegraphics[width=3.6cm]{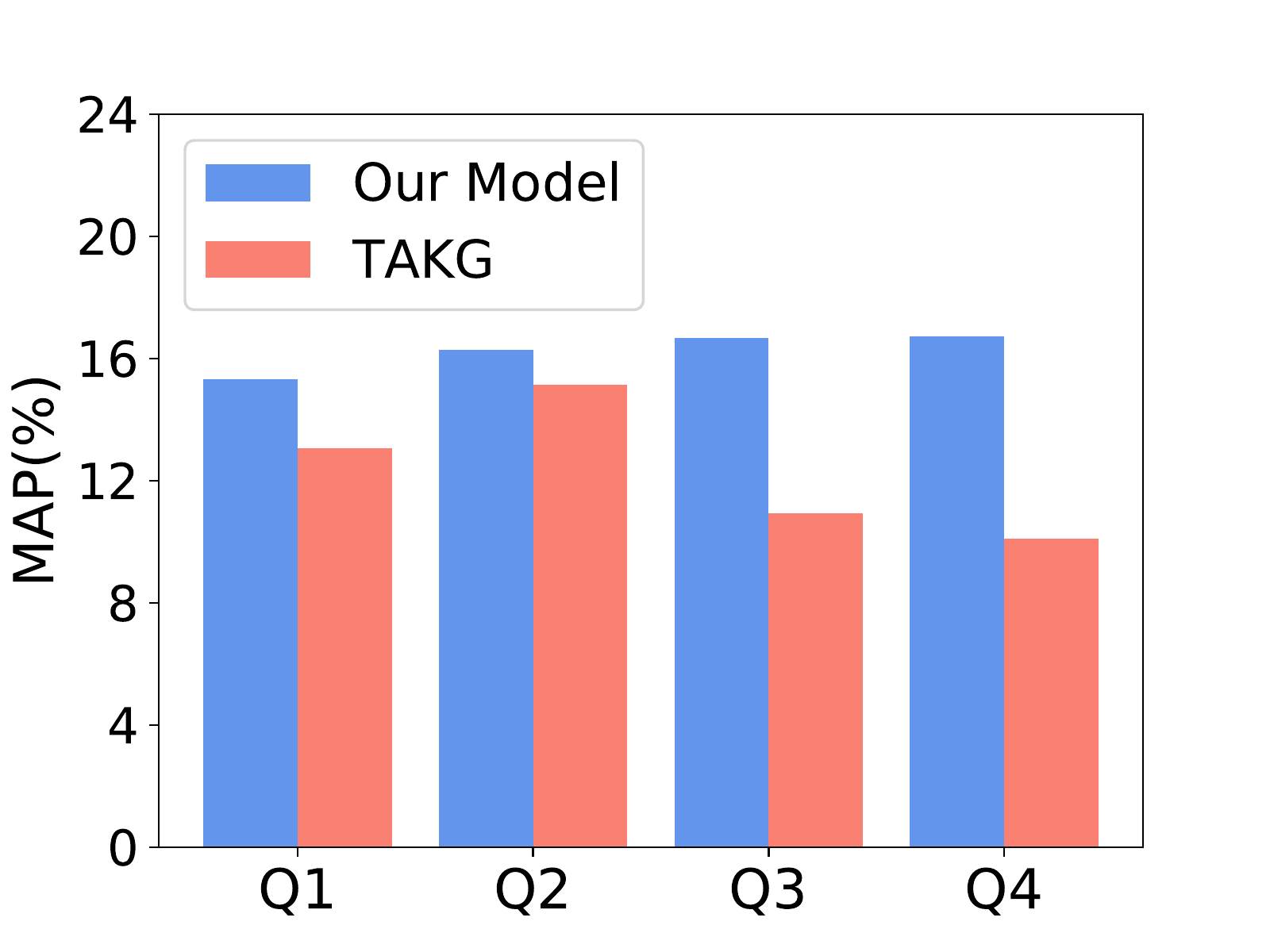}
}
\caption{
\label{fig:turn_len}
%Recommendation Results on different turn length and context length.
MAP scores (y-axis) over context length (left in turn number and right token number) in varying quantiles. For each subfigure, from left to right shows the results in $Q_1$ ($[0, 0.25)$), $Q_2$ ($[0.25, 0.5)$), $Q_3$ ($[0.5, 0.75)$), and $Q_4$ ($0.75, 1)$).
}
\end{figure}

To further explore model's sensitivity to turn number% in both training and test process
, we first rank the conversations with turn number and separate them into four quartiles ($Q_1,Q_2,Q_3,Q_4$, in order with increasing turn number).
We then train and test in each quartile, and compare the results of our model and \textsc{TAKG} in Figure~\ref{sfig:turn}. 
As can be seen, our model presents larger margin for quartiles corresponding to larger turn number, indicating our ability to encode rich information from complex turn interactions. 

%The performance of our model is more stable, while \textsc{TAKG} does decline significantly.

%To further emphasize the robustness of our model, we re-train and test our model based on conversations with only certain turn numbers, where we divide the Reddit dataset into $4$ parts, each with same level of turn number. The results, together with model \textsc{TAKG}'s, are displayed in Figure~\ref{sfig:turn}.
%In order to further verify our above point of view, we try to our model's performance in datasets with different average turn length. We sort Reddit dataset according to the length of turn, and then cut them into 4 sub datasets with equal size. Then we do training, validation and testing on those 4 sub dataset respectively, and the results are in Figure \ref{sfig:turn}. The best baseline TAKG's results are also compared.
% As can be seen, the performance gap between our model and \textsc{TAKG} increases as the average turn number increases. The performance of our model is more stable, while \textsc{TAKG} does decline significantly.

 \textit{Token number.} For context length measured with token number, we follow the above steps to form train and test quartiles for token number. The results are shown in Figure \ref{sfig:len} where our model consistently outperform TAKG over conversation context with varying token number.
%We also want to examine the effects of token number as context. Similar to what we do for turn number, we split the Reddit dataset into $4$ parts with different average token numbers, and also re-train and test.
%Because the length of turn is related to the length of context to a certain extent, in order to ensure that the performance of our model on turn is not caused by the length of context, we further explore the impact of context length on model performance. We sort Reddit dataset according to the length of context (including query), and then cut them into 4 sub datasets with equal size. Then we do training, validation and testing on those 4 sub datasets respectively.
%The results are in Figure \ref{sfig:len}. We can see that \textsc{TAKG} can perform quite well with a certain number of tokens (as shown in LEN2), and then decrease. But our model performs consistently better in all different numbers of tokens, again indicating our robustness.

\paragraph{The Effects of Quotation.}
%Different kinds of quotations may exhibit differently when recommending, especially for English quotes, whose patterns are more varied.

%\noindent \textit{Quotation frequency.}

\begin{figure}[t]
\centering
\subfigure[MAP on Freq. Ranges]{\label{sfig:fre_left}
\includegraphics[width=3.6cm]{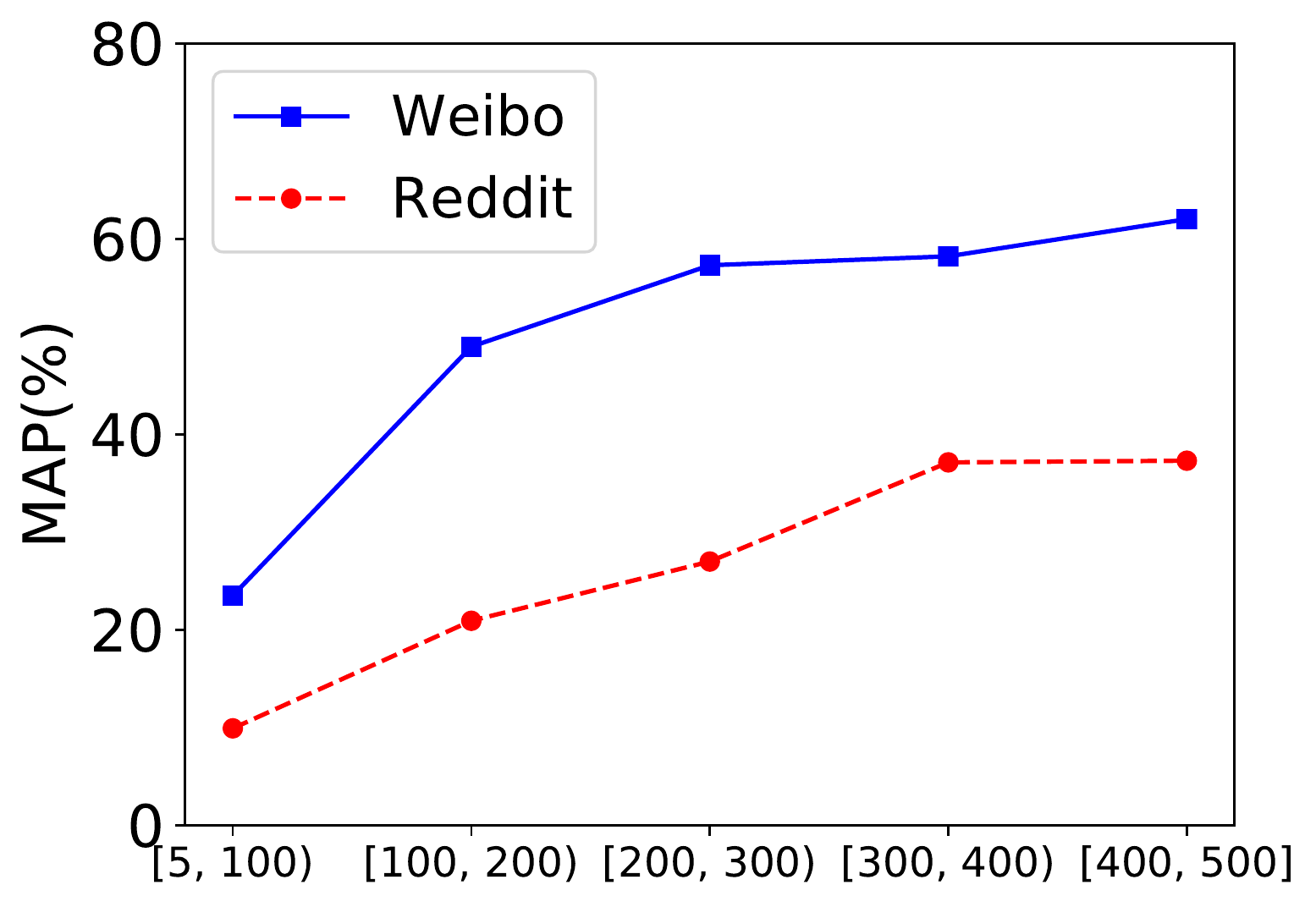}
}
\subfigure[Growth Rate Comparison]{\label{sfig:fre_right}
\includegraphics[width=3.6cm]{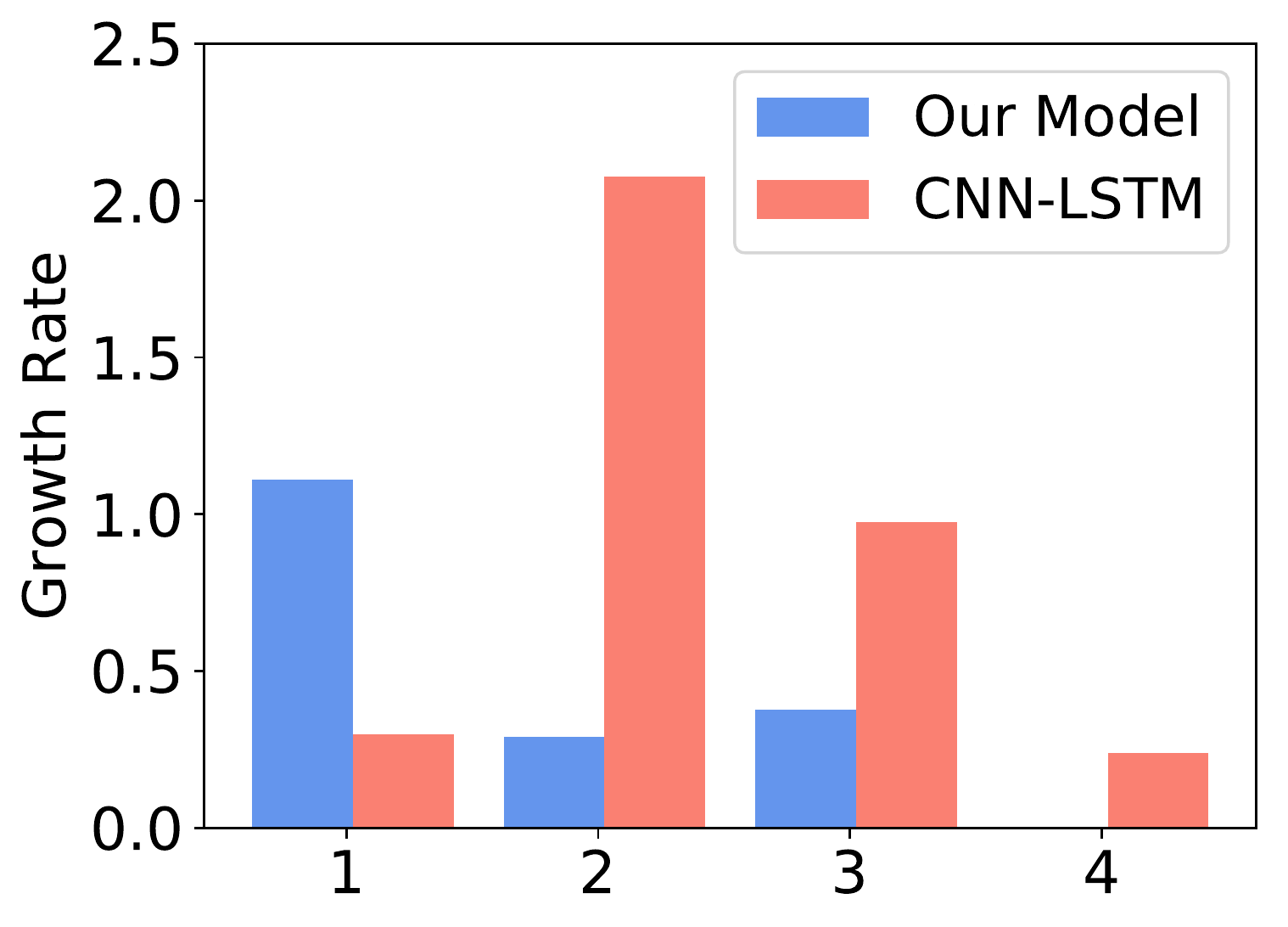}
}
\caption{
\label{fig:fre_map}
%Recommendation Results on different turn length and context length.
Left subfigure: Our MAP scores (y-axis) on different frequency range (x-axis). 
Right: growing rate (y-axis) on Weibo, where x-axis indicates the order of neighboring ranges.
}
\end{figure}

% \includegraphics[width=0.32\textwidth]{pictures/fre_map.pdf}
% \vskip -0.5em
% \caption{\label{fig:fre_map} Our model's performance on different frequency ranges. X-axis indicates the quotation's frequency range.} 
% \vskip -1em
% \end{figure}

We further study our results to predict quotations in varying frequency and the MAP scores are reported in Figure~\ref{sfig:fre_left}.
In general, higher scores are observed for more frequent quotations, as better representations can be extensively learned from training data. 
We also notice a slower growing rate as the frequency increases.
To go into more details, we compare the growing rates with ranking model \textsc{CNN+LSTM} and show the results in \ref{sfig:fre_right} on Weibo (Reddit results in similar trends).
In comparison, we are generally less sensitive to quotation frequency (except for very rare quotes). It is likely to be benefited from quotations' internal structure while ranking models can be largely affected by label sparsity.

%We also compare the MAP's growth rate between ranges with ranking model \textsc{CNN-LSTM} and find that \textsc{CNN-LSTM} is more sensitive to data sparsity as shown in \ref{sfig:fre_right}.

%In addition to quotation frequency, we also study our sensitivity to quotation length and find that our model is not sensitive to such factor. 
%It is probably because quotations exhibit relatively static patterns.

%Some quotations may be more familiar to people, so appear more frequently on online discussions. We would like to know whether different appearing frequencies will result in different performance.
%Figure~\ref{fig:fre_map} shows our performance towards quotations with different frequencies. 
%As it shows, the more frequent a quotation is, the better performance will be.

% We also examine the effects of quotation length  find that our model is not sensitive to such factor.
% We are curious whether the length of quotations will affect the difficulty of recommendation. We conduct statistics on the performance with different length of quotations. Then we find that the performance varies as the quote length increases, but is not simple linear relationship.
% Therefore, the length of quotation indeed affects the performance, but are of more complex relation.

\subsection{Further Discussions}\label{ssec:res:further}

Here we probe into our outputs to provide more insights to quoting in conversations.

%We further dive into the result details for our model in this section, to analyze the superiority of our model.

\paragraph{Case Study.}
\begin{figure}[t]
\centering
\includegraphics[width=0.45\textwidth]{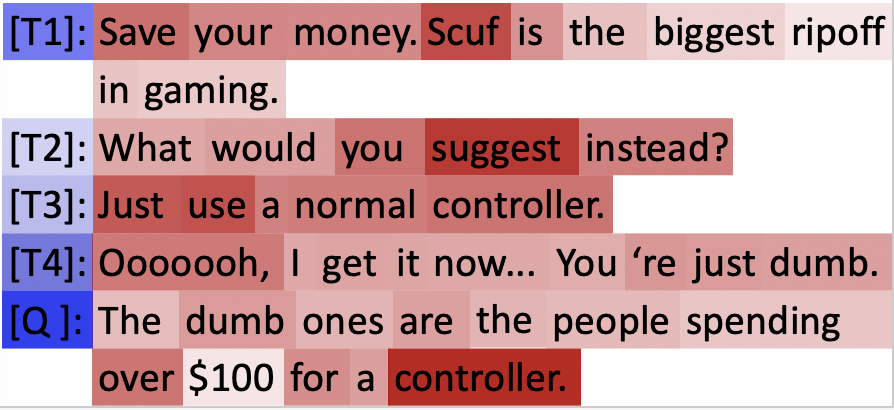}
\caption{\label{fig:case-attention} Attention weights of our model over turns (in blue) and \textsc{TAKG} over words (in red).}
\end{figure}

We first present a qualitative analysis over the example in 
%We take the case conversation  
Figure \ref{fig:intro-example}. To analyze what the model learns, we visualize our turn-based attention and \textsc{TAKG}'s topic-aware attention over words in Figure \ref{fig:case-attention}.
As can be seen, \textsc{TAKG} focuses more on topic words ``Scuf'', ``suggest'', and ``controller'', all reflecting the global discussion focus while ignoring query's intention.
Thus, it mistakenly quote ``\textit{A penny saved is a penny earned.}''.
Instead, we attend the query's interaction with $T_1$ and $T_4$, which results in the correct quotation. 

\paragraph{Comparing with Human.}
Finally, we discuss how human performs on our task. $50$ Weibo conversations were hence sampled and two human annotators (native Chinese speakers) were invited to quote a Chinese Chengyu in the given context. 
The two annotators give $7$ and $8$ correct answers respectively, which shows the task is challenging for human.
Our model made $13$ correct predictions, exhibiting a better ability to quote in online conversations.
%We are also interested in whether it is a challenging task for real humans to predict and recommend quotations.
%Therefore, we set up an human evaluation as below. 
%We randomly sample 50 conversations (end with query turns without final quotations) from Weibo dataset, and invite 2 native Chinese speakers to complete the conversations with idioms.
%The two humans get 7, 8 correct idioms respectively, which is very poor and not comparable to the performance of our model. 

%We are also interested in how human performs in such quotation recommendation task, in order to find out how challenge this task is. We randomly sample 50 conversations from Weibo dataset and invite 2 native Chinese speakers to complete the conversations with idioms. The two humans get 7, 8 correct idioms respectively, which is not comparable to the performance of our model. 

\section{Conclusion}

We present a novel quotation generation framework for online conversations via the modeling of topic, interaction, and query consistency.
Experiment results on two newly constructed online conversation datasets, Weibo and Reddit, show that our model outperforms the previous state-of-the-art models. Further discussions provide more insights on quoting in online conversations.

%We proposed a novel quotation generation model that can keep the language consistency between quotations and conversations. Experimental results on two newly constructed online conversation datasets show that our model outperforms the state-of-the-art quote recommendation models. 
%Further discussion shows that our model performs consistently better regarding different context and quotations, indicating the superiority and effectiveness of our model.
%Further discussion shows that our model's design considers the characteristic of online conversations and effective. 
%In the future, we will explore how to generate quotable sentences not only limited to quotations. Also, we can explore why quotations are quotable in a linguistic perspective. 

\section*{Acknowledgements}
The research described in this paper is partially supported by CUHK RSFS \#3133237, HK RGC GRF \#14204118, HK ITF \#ITS/335/18.
Jing Li is supported by the Hong Kong Polytechnic University internal funds (1-BE2W and 1-ZVRH) and NSFC Young Scientists Fund 62006203.
\bibliographystyle{acl_natbib}
\bibliography{acl2020}

\end{document}